\documentclass{article} 
\usepackage{nips11submit_e,times}

\usepackage{amsmath}
\usepackage{amsfonts}
\usepackage{amssymb}
\usepackage{amsthm}
\usepackage{graphicx}

\newtheorem{theorem}{Theorem}
\newtheorem{definition}{Definition}
\newtheorem{lemma}{Lemma}

\newcommand{\given}{\, | \,}
\newcommand{\Prob}{\mathbf{P}}
\newcommand{\set}[1]{\mathcal{#1}}

\renewcommand{\vec}[1]{\boldsymbol{#1}}
\newcommand{\class}{y}
\newcommand{\on}[1]{\operatorname{#1}}
\newcommand{\nlab}{M}
\newcommand{\ninst}{N}
\newcommand{\fromto}{\longrightarrow}
\newcommand{\aset}{\set{Y}}
\newcommand{\lab}{y}

\title{Label Ranking with Abstention: Predicting Partial Orders by Thresholding Probability Distributions (Extended Abstract)}

\author{
Weiwei Cheng\\
Mathematics and Computer Science\\
University of Marburg, Germany\\
\texttt{cheng@mathematik.uni-marburg.de}\\
\And
Eyke H\"ullermeier \\
Mathematics and Computer Science\\
University of Marburg, Germany\\
\texttt{eyke@mathematik.uni-marburg.de}\\
}

\nipsfinalcopy 

\begin{document}

\maketitle

\begin{abstract}
We consider an extension of the setting of label ranking, in which the learner is allowed to make predictions in the form of partial instead of total orders. Predictions of that kind are interpreted as a partial abstention: If the learner is not sufficiently certain regarding the relative order of two alternatives, it may abstain from this decision and instead declare these alternatives as being incomparable. We propose a new method for learning to predict partial orders that improves on an existing approach, both theoretically and empirically. Our method is based on the idea of thresholding the probabilities of pairwise preferences between labels as induced by a predicted (parameterized) probability distribution on the set of all rankings.
\end{abstract}

\section{Introduction}

In the setting of label ranking, a special type of preference learning problem,
each instance $\vec{x}$ from an instance space $\mathbb{X}$  
is associated with a total order of a fixed set of class labels $\set{Y} = \{ y_1, \ldots , y_M\}$, that is, a complete, transitive, and asymmetric relation $\succ_{\vec{x}}$ on $\set{Y}$, where $\class_i \succ_{\vec{x}} \class_j$ indicates that, for instance $\vec{x}$, $\class_i$ precedes $\class_j$ in the order. 
Since a ranking can be considered as a special type of preference relation, we shall also say that $\class_i \succ_{\vec{x}} \class_j$ indicates that $\class_i$ is \emph{preferred} to $\class_j$ given the instance $\vec{x}$. 

Formally, a total order $\succ_{\vec{x}}$ can be identified with a permutation $\pi_{\vec{x}}$ of the set $\{1 , \ldots , \nlab \}$, such that $\pi_{\vec{x}}(i)$ is the index $j$ of the class label $y_j$ on the $i$-th position in the order (and hence $\pi^{-1}_{\vec{x}}(j) = i$ the position of the $j$-th label). This permutation thus encodes the (ground truth) order relation
\begin{equation*}
\class_{\pi_{{\boldsymbol{x}}}(1)} \succ_{{\boldsymbol{x}}} \class_{\pi_{{\boldsymbol{x}}}(2)} \succ_{{\boldsymbol{x}}} \ldots
\succ_{{\vec{x}}} \class_{\pi_{\vec{x}}(\nlab)} \enspace .
\end{equation*}
We denote the class of permutations of $\{1 , \ldots , \nlab \}$ (the symmetric group of order $\nlab$) by $\Omega$. 

The goal in label ranking is to learn a ``label ranker'' in the form of an $\mathbb{X} \fromto \Omega$ mapping. As training data, a label ranker uses a set of instances ${\vec{x}}_n$ ($n=1,  \ldots , \ninst$), together with preference information 
in the form of pairwise comparisons $\class_i \succ_{\vec{x}_n} \class_j$ of some labels in $\set{Y}$, suggesting that instance $\vec{x}_n$ prefers label $\class_i$ to $\class_j$. 

Motivated by the idea of a reject option in classification, the authors in \cite{mpub205} introduced a variant of the above setting in which the label ranker is allowed to partially abstain from a prediction. More specifically, it is allowed to make predictions in the form of \emph{partial} instead of \emph{total} orders: If the ranker is not sufficiently certain regarding the relative order of two alternatives and, therefore, cannot reliably decide whether the former should precede the latter or the other way around, it may abstain from this decision and instead declare these alternatives as being incomparable. Abstaining in a consistent way, it should of course still produce an asymmetric and transitive relation, hence a partial order.

The approach in \cite{mpub205}, despite being the first to address the problem of learning to predict partial orders, still exhibits some disadvantages (see next section). In this paper, we therefore propose an alternative method, or rather a modification, which is based on the idea of predicting partial orders by thresholding parameterized probability distributions on rankings. Roughly speaking, by making stronger model assumptions, this approach is able to avoid inconsistencies that may occur in \cite{mpub205}, and hence simplifies the construction of consistent partial order relations; see Section 3 for details. 

Of course, despite being interesting from a theoretical point of view, these properties do not guarantee a practical advantage in terms of prediction performance, especially in cases where the model assumptions might be violated. Therefore, we complement our theoretical results by an experimental study in which we compare our new method with the original approach of \cite{mpub205}.

\section{Previous Work}

The method in \cite{mpub205} consists of two main steps and can be considered as a pairwise approach in the sense that, as a point of departure, a valued preference relation $P: \aset \times \aset \rightarrow [0,1]$ is produced, where $P(\lab_i,\lab_j)$ is interpreted as a measure of support of the pairwise preference $\lab_i \succ \lab_j$. Support is commonly interpreted in terms of probability, hence $P$ is assumed to be reciprocal, that is, $P(\lab_i,\lab_j) \, = \, 1 - P(\lab_j,\lab_i)$ for all $\lab_i,\lab_j \in \aset$. Then, in a second step, a partial order $Q$ is derived from $P$ via thresholding: $Q(\lab_i,\lab_j) = 1$ if  $P(\lab_i,\lab_j) > q$ and  $Q(\lab_i,\lab_j) = 0$ otherwise, where $1/2 \leq q < 1$ is a threshold. Thus, the idea is to predict only those pairwise preferences that are sufficiently likely, while abstaining on pairs $(\lab_i,\lab_j)$ for which the probability $P(\lab_i,\lab_j)$ is too close to $1/2$.

The first step of deriving the relation $P$ is realized in \cite{mpub205} by means of an ensemble learning technique: Training an ensemble of standard label rankers, each of which provides a prediction in the form of a total order, $P(\lab_i,\lab_j)$ is defined by the fraction of ensemble members voting for $\lab_i \succ \lab_j$. Other possibilities are of course conceivable, and indeed, the only important point to notice here is that the preference degrees $P(\lab_i,\lab_j)$ are essentially independent of each other. Or, stated differently, they do not guarantee any specific properties of the relation $P$ except being reciprocal. For the relation $Q$ derived from $P$ via thresholding, this has two important consequences:
\begin{itemize}
\item If the threshold $q$ is not large enough, then $Q$ may have cycles. Thus, not all thresholds in $[0.5, 1)$ are actually feasible. In particular, if $q=0.5$ cannot be chosen, this also implies that the method may not be able to predict a total order as a special case.
\item Even if $Q$ does not have cycles, it is not guaranteed to be transitive.
\end{itemize}
To overcome these problems, the authors devise an algorithm that finds the smallest feasible threshold $q_{min}$ and ``repairs'' a non-transitive relation $Q$ by replacing it with its transitive closure. The complexity of this algorithm is $\mathcal{O}(|\aset|^3)$.

\section{Predicting Partial Orders based on Probabilistic Models}

In order to tackle the above problems, our idea is to restrict the relation $P$ so as to exclude the possibility of cycles and violations of transitivity  from the very beginning. To this end, we take advantage of methods for label ranking that produce (parameterized) \emph{probability distributions} over $\Omega$ as predictions. Our main theoretical result is to show that thresholding pairwise preferences induced by such distributions yields preference relations with the desired properties, that is, partial order relations $Q$. 

In \cite{mpub192}, a label ranking method was proposed that produces predictions expressed in terms of the Mallows model \cite{mard_aa}, a \emph{distance-based} probability model belonging to the family of exponential distributions. The standard Mallows model 
\begin{equation}\label{eq:mallows}
\Prob(\pi \given \theta,\pi_0) = \frac{\exp(- \theta D(\pi,\pi_0))}{\phi(\theta)} 
\end{equation}
is determined by two parameters:
The ranking $\pi_0 \in \Omega$  is the location parameter (mode, center ranking) and $\theta \geq 0$ is a spread parameter. Moreover, $D$ is a distance measure on rankings, and the constant $\phi = \phi(\theta)$ is a normalization factor that depends on the spread (but, provided the right-invariance of $D$, not on $\pi_0$). 
Obviously, the Mallows model assigns the maximum probability to the center ranking $\pi_0$. The larger the distance $D(\pi,\pi_0)$, the smaller the probability of $\pi$ becomes. The spread parameter $\theta$ determines how quickly the probability decreases, i.e., how peaked the distribution is around $\pi_0$. For $\theta = 0$, the uniform distribution is obtained, while for $\theta \rightarrow  \infty$, the distribution converges to the one-point distribution that assigns probability 1 to $\pi_0$ and 0 to all other rankings.

Alternatively, the Plackett-Luce (PL) model was used in \cite{mpub208}. This is a stagewise model, which is specified by a parameter vector $\vec{v} = (v_1, v_2, \ldots , v_\nlab) \in \mathbb{R}_+^\nlab$ \cite{mard_aa}:
\begin{equation}\label{eq:pl}
\Prob(\pi \given \vec{v}) \, = \, \prod_{i=1}^\nlab \frac{v_{\pi(i)}}{v_{\pi(i)} + v_{\pi(i+1)} + \ldots + v_{\pi(\nlab)}} 
\end{equation}
This model is a generalization of the well-known Bradley-Terry model for the pairwise comparison of alternatives, which specifies the probability that ``$a$ wins  against $b$'' in terms of $\Prob(a \succ b) \, = \, \frac{v_a}{v_a + v_b}$. 
Obviously, the larger $v_a$ in comparison to $v_b$, the higher the probability that $a$ is chosen. Likewise, the larger the parameter $v_i$ in (\ref{eq:pl}) in comparison to the parameters $v_j$, $j \neq i$, the higher the probability that the label $y_i$ appears on a top rank. An intuitively appealing explanation of the PL model can be given in terms of a vase model: If $v_i$ corresponds to the relative frequency of the $i$-th label in a vase filled with labeled balls, then $\Prob(\pi \given \vec{v})$ is the probability to produce the ranking $\pi$ by randomly drawing balls from the vase in a sequential way and putting the label drawn in the $k$-th trial on position $k$ (unless the label was already chosen before, in which case the trial is annulled).

Given a probability distribution $\Prob$ on the set of rankings $\Omega$, the probability of a pairwise preference $\class_i \succ \class_j$ (and hence the corresponding entry in the preference relation $P$) can be  derived through marginalization: 
\begin{equation}\label{acc}
P( \class_i  , \class_j) = \Prob(\class_i \succ \class_j)  \, = \, \sum_{\pi  \in E(\class_i , \class_j)} \Prob(\pi ) 
\enspace ,
\end{equation}
where $E(\class_i , \class_j)$ denotes the set of linear extensions of the incomplete ranking $\class_i \succ \class_j$, i.e., the set of all rankings $\pi \in \Omega$ in which $\lab_i$ precedes $\lab_j$. Our main theoretical result states that thresholding (\ref{acc}) yields a proper partial order relation $Q$, both for the Mallows and the PL model. 

\begin{theorem}\label{theorem}
Let $\Prob$ in (\ref{acc}) be the Mallows model (\ref{eq:mallows}), with a distance $D$  having the so-called transposition property, or the PL model (\ref{eq:pl}). Moreover, let $Q$ be defined by the thresholded relation $Q(\lab_i,\lab_j)= 1$ if $P(\lab_i,\lab_j) > q$ and $Q(\lab_i,\lab_j)= 0$ otherwise. Then $Q$ defines a proper partial order relation for all $q \in [1/2, 1)$. 
\end{theorem}

A distance $D$ on rankings is said to have the \emph{transposition property}, if the following holds: Let $\pi$ and $\pi'$ be rankings so that, in both of them, $y_i$ precedes $y_j$. Moreover, consider a third ranking $\pi''$ identical to $\pi'$, except for a transposition of $y_i$ and $y_j$. Then, $D(\pi, \pi') \leq D(\pi, \pi'')$. Of course, this property is intuitively plausible, and indeed, it is satisfied by most of the commonly used distance measures (see, e.g., \cite{critchlow91}).

While the proof of the above theorem is rather straightforward for the PL model, it becomes less obvious in the case of the Mallows model. In any case, it guarantees that a proper partial order relation can be predicted by simple thresholding, and without the need for any further reparation. Moreover, the whole spectrum of threshold parameters $q \in [1/2, 1)$ can be used.

\section{Experiments}

As mentioned earlier, the alternative approach outlined above does not automatically imply a practical advantage, especially since it makes strong model assumptions (in terms of the Mallows or PL model) that are not necessarily satisfied. Therefore, we complement our theoretical results by an empirical study, in which we analyze the tradeoff between \emph{correctness} and \emph{completeness} achieved by different methods.

If a model is allowed to abstain from making predictions, it is expected to reduce its error rate. In fact, it can trivially do so, namely by rejecting all predictions, in which case it avoids any mistake. Clearly, this is not a desirable solution. Indeed, in the setting of prediction with reject option, there is always a trade-off between two criteria: \emph{correctness} on the one side and \emph{completeness} on the other side. An ideal learner is correct in the sense of making few mistakes, but also complete in the sense of abstaining  rarely. The two criteria are conflicting: increasing completeness typically comes along with reducing correctness and vice versa, at least if the learner is effective in the sense that it abstains from those decisions that are indeed most uncertain. 

As measures of correctness and completeness, we use those that were proposed in \cite{mpub205}. Correctness is measured by the \emph{gamma rank correlation} (between the true ranking and the predicted partial order), and completeness is defined by one minus the (relative) number of pairwise comparisons on which the model abstains.

\begin{figure}
\begin{center}
\includegraphics[width=0.8\textwidth]{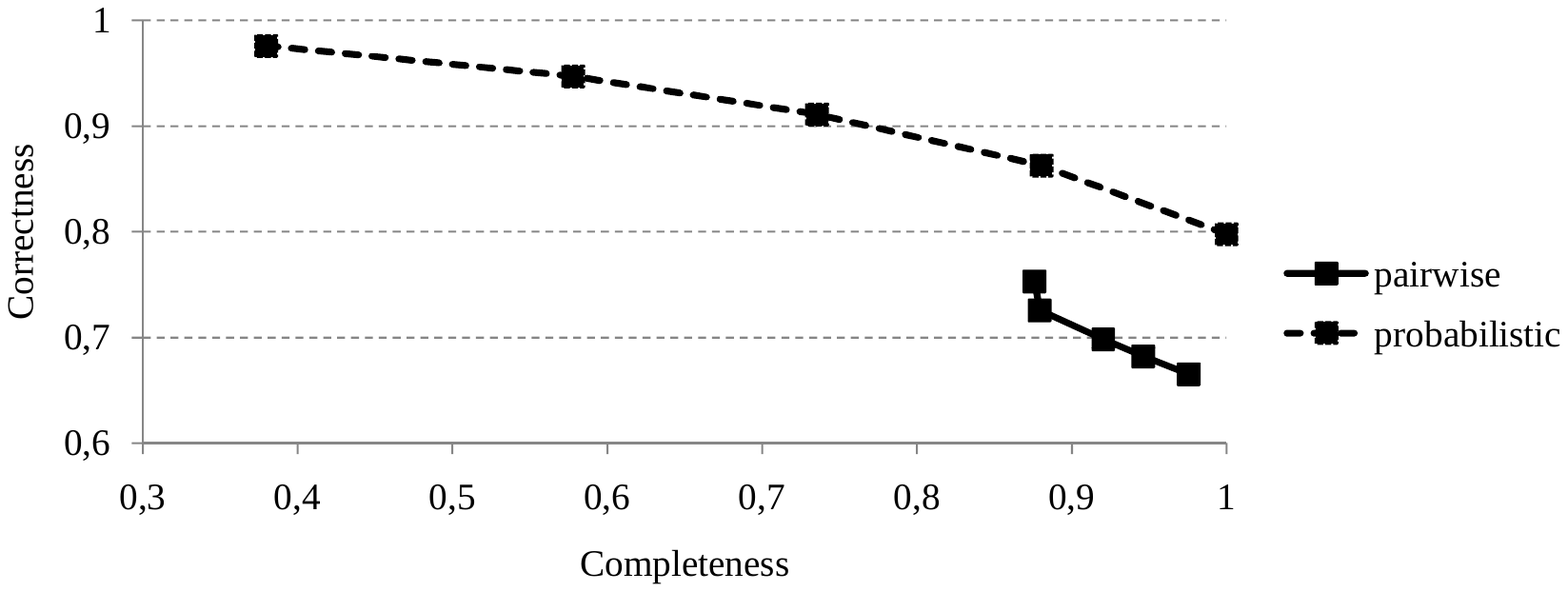} 
\end{center}
\caption{Trade-off between completeness and correctness for a label ranking variant of the UCI benchmark data set VOWEL: Existing pairwise method (solid line) versus new approach based on probabilistic models (dashed line).}
\label{fig:results}
\end{figure}

The main conclusion that can be drawn from our results is that, as expected, our probabilistic approach does indeed achieve a better trade-off between completeness and correctness, especially in the sense that it spans a wider range of values for the former. Besides, we often observe that the level of correctness is increased, too. A typical example of the completeness/complexity trade-off is shown in Figure 1.

\bibliography{lit}
\bibliographystyle{plain}

\end{document}